\documentclass{article} 
\usepackage{iclr2021_conference,times}

\usepackage{amsmath,amsfonts,bm}

\def\eqref#1{equation~\ref{#1}}

\def\1{\bm{1}}

\DeclareMathAlphabet{\mathsfit}{\encodingdefault}{\sfdefault}{m}{sl}
\SetMathAlphabet{\mathsfit}{bold}{\encodingdefault}{\sfdefault}{bx}{n}

\usepackage{bbm}
\usepackage{lipsum}
\usepackage{amsmath}
\usepackage{amssymb}
\usepackage{physics}
\usepackage{commath}
\usepackage{xspace}
\usepackage{microtype}
\usepackage{tikz}
\usepackage{pgf}
\usepackage{color}
\usepackage{import}
\usepackage{comment}
\usepackage{multirow}
\usepackage{wrapfig}
\usepackage{booktabs}
\usepackage{pgfplots}
\usepackage[english]{babel}
\usepackage[hyphens]{url}
\usepackage{subcaption}
\usepackage{listings}

\newcommand{\twofourtwo}{\textsc{Two4Two}\xspace}

\title{Two4Two: Evaluating Interpretable Machine Learning -- A Synthetic Dataset For Controlled Experiments}

\def\weiz{\textsuperscript{1}}
\def\fu{\textsuperscript{2}}
\author{%
Martin Schuessler\weiz, Philipp Wei\ss{}\weiz, Leon Sixt\fu  \\
schuessler@tu-berlin.de, leon.sixt@fu-berlin.de, philipp@itp.tu-berlin.de \\
\weiz Weizenbaum Institut, TU Berlin \fu Freie Universität Berlin
}

\iclrfinalcopy 
\begin{document}

\maketitle

\begin{abstract}
A growing number of approaches exist to generate explanations for image classification.
However, few of these approaches are subjected to human-subject evaluations, partly because it is challenging to design controlled experiments with natural image datasets, as they leave essential factors out of the researcher's control.
With our approach, researchers can describe their desired dataset with only a few parameters. Based on these, our library generates synthetic image data of two 3D abstract animals. The resulting data is suitable for algorithmic as well as human-subject evaluations. 
Our user study results demonstrate that our method can create biases predictive enough for a classifier and subtle enough to be noticeable only to every second participant inspecting the data visually.
Our approach significantly lowers the barrier for conducting human subject evaluations, thereby facilitating more rigorous investigations into interpretable machine learning.\footnote{%
Our library and datasets are available at \url{https://github.com/mschuessler/two4two/}}
\end{abstract}

\section{Introduction}
Researchers are faced with an abundance of approaches to generate explanations for image classification and segmentation. However, for many such methods, it remains unclear if they explain a model faithfully \citep{adebayo2018sanity,sixt2020wel,leavitt2020towards,nie2018theoretical} and if they provide any utility to users. Just recently, researchers began to evaluate explanations algorithmically on synthetic benchmark datasets, using ground truth segmentation data to validate feature attributions.
The CLEVR-XAI dataset \citep{arras2021} provides such ground truth for visual question answering while
the BAM dataset \citep{Yang2019} focuses on image classification.

However, algorithmic evaluations are insufficient as intelligibility is a human-centered concept.
Recent human subject evaluations raise concerns about the claimed utility of explanations to facilitate model understanding \citep{Alqaraawi2020}, trust calibration \citep{Kaur2020,Chu2020}, and error understanding \citep{Shen2020}.  
Such results emphasize that claims about intelligibility 
require human subject experiments to be validated
 \citep{Doshi2017}.
Up until now very few such evaluations are conducted. Probably because they are a challenging endeavor, requiring to mimic a realistic setting while avoiding overburdening participants \citep{Doshi2017,wortmanvaughan2021a,Adadi2018,Nunes2017}.

Researchers first need to make some deliberate design choices regarding the used dataset to conduct human grounded evaluations.
In this work, we introduce a synthetic data generation library specifically designed for this task.

\section{Synthetic Image Data For Interpretabiltiy Evaluations}

There are many natural image datasets for image classification and segmentation. Domain-specific datasets, such as medical imagery, are only meaningful to experts whose recruitment is often infeasible. Hence, a more economical and practical approach is to recruit lay participants through crowdsourcing platforms. This further limits the choice of natural image datasets to everyday imagery. Popular default datasets appear to be MNIST, ImageNet, and PASCAL. 
Researchers usually consider only a fraction of the available classes to limit task complexity and study duration. 
Animal labels seem to be a prevalent choice (e.g. \cite{Ribeiro2016,Kim2016,Adebayo2020,Goyal2019,Alqaraawi2020}. 
While often perfectly capable of recognizing these labels, participants might have different preconceptions of how a machine learning system should detect them.
For example, \cite{Alqaraawi2020} noted that some participants began reasoning about specific horse-riding equipment, indicating that some participants had more domain knowledge than others.
Hence, natural images introduce participants' prior knowledge  as a confounding factor. 

A user study's straightforward and reasonable goal is to evaluate if participants can, with the help of an explanation technique, discover the features and biases a model uses for its predictions.
However, this requires ground truth information about feature importance \citep{Yang2019}.
For example, to test whether participants identify an image's background 
as a predictive feature for a given label, the background needs to be an independently controllable factor.

The BAM dataset by \cite{Yang2019} is a controllable data generation module. It overlays a foreground object on top of a natural background image. One can use it to deliberately bias a model by introducing correlations between the foreground and background. However, the resulting images appear artificial, eliminating the single major advantages of natural images for user studies. After all, a dog placed randomly in a bamboo forest might confuse participants. To circumvent this problem, we choose to create a synthetic dataset. 
\cite{arras2021} also use synthetic data in the  CLEVR-XAI  dataset. 
However, this dataset focuses on algorithmically evaluating attribution methods, testing if the correct object is highlighted. Adding spurious correlation is not intended.
We concluded that a dataset suited for human evaluation with independently controllable biases is currently missing.

\section{\twofourtwo: Data Generation and Parameter Sampling}

\begin{figure}
    \vspace{-1.2cm}
    \centering
    \begin{subfigure}[b]{0.48\textwidth} 
        \hspace{2cm}
        \includegraphics[height=3.2cm]{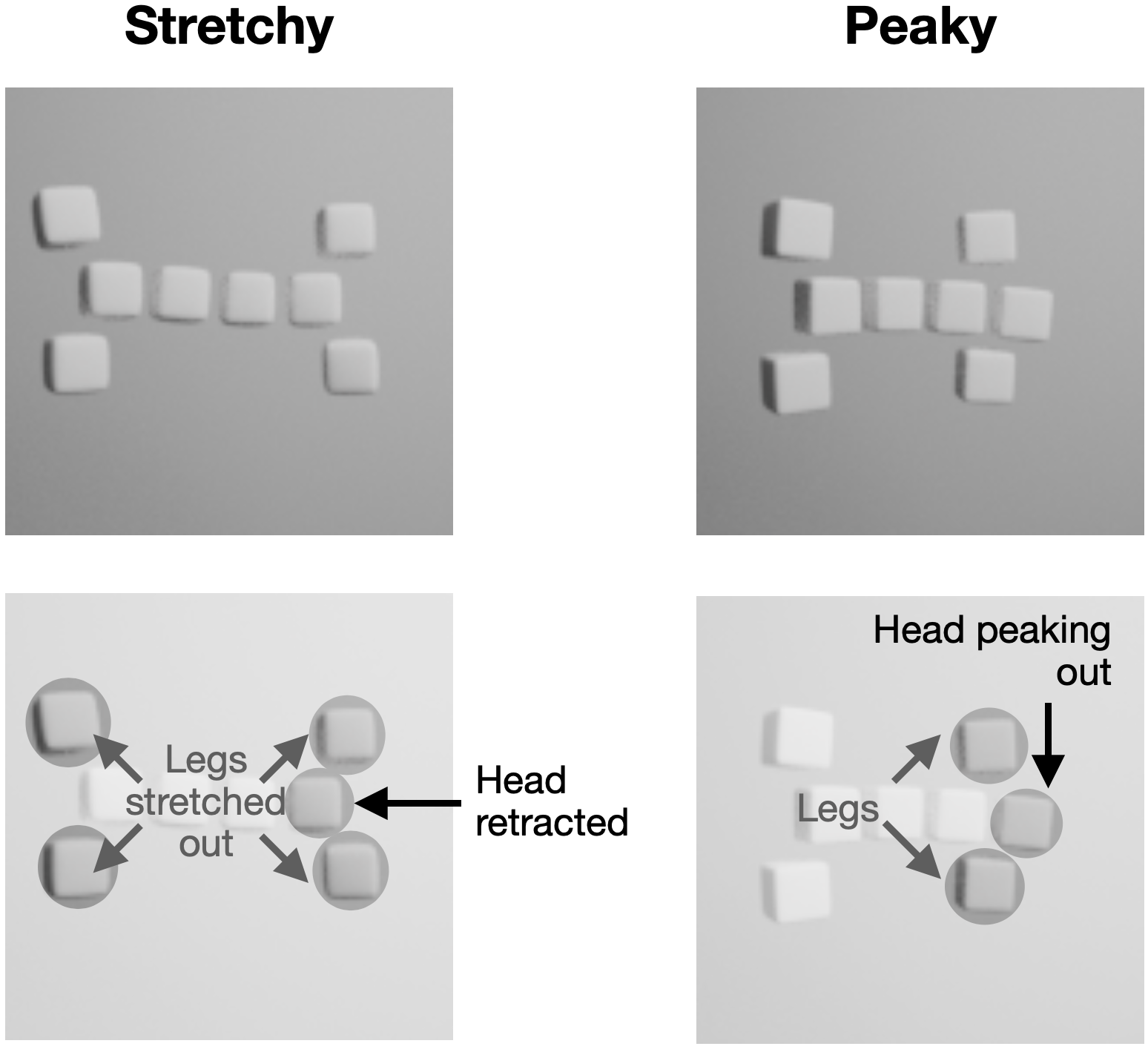}
        \caption{Peaky and Stretchy}
        \label{fig:peaky_and_stretch}
    \end{subfigure}
    \begin{subfigure}[b]{0.48\textwidth} 
        \includegraphics[height=3.2cm]{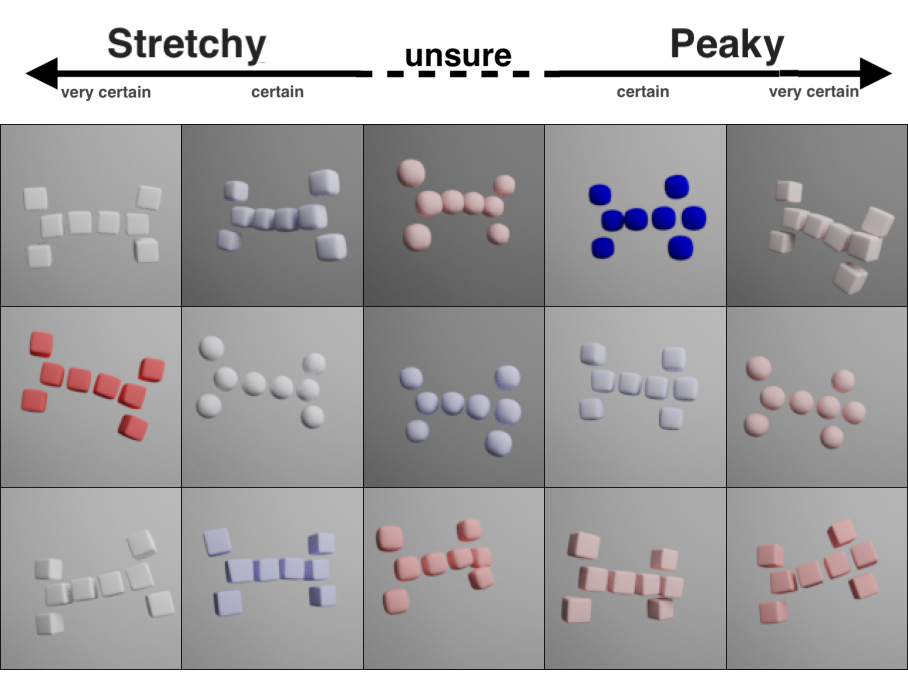} 
        \caption{Interface of the user study}
        \label{fig:user_study_interface}
    \end{subfigure}
    \vspace{-0pt}
    \caption{
    \textbf{(a)} Objects in the \twofourtwo dataset:  \emph{Peaky}
    and \emph{Stretchy}.
    Each animal consists of a spine of four blocks and two sets of arms at either end. 
    \emph{Peaky}, has the right set of arms moved inwards. \emph{Stretchy} class, has both sets of arms are moved outwards.
    \textbf{(b)} A crop of the 30$\times$5 grid ordered by classification score. It was used as example based baseline explanation technique in the user study.    }
    \label{fig:peaky_and_interface}
    \vspace{-0.5cm}
\end{figure}

We developed a library that allows researchers to render synthetic image data suitable for human-subject evaluations. 
The library can be used for algorithmic evaluation as well. The generated segmentation masks allow 
comparing relevant image areas as done for the CLEVR-XAI dataset \citep{arras2021}. Introducing different biases as in the BAM dataset \cite{Yang2019} is a core functionality of our dataset. Although we believe that \twofourtwo is a 
good fit for algorithmic evaluation, we leave it to
future work.

Our dataset has only two classes: the abstract ``animals'' named \emph{Peaky} (arms inwards) and \emph{Stretchy} (arms outwards).
Each comprises eight building blocks, of which four blocks in a row make up the ``spine'' of both animals. At each end of the spine, two additional blocks are attached above and below. They constitute the ``arms''. 
Two arms, 4 spines, and two arms -- \twofourtwo (See Figure \ref{fig:peaky_and_stretch}).
This composition is simple enough to be used in instructions for human-subject evaluations. 
Study participants can differentiate the two classes by looking at the arm position relative to the spine.
How an animal is portrayed within a scene is described by a small set of mutable parameters:
\begin{itemize}

\item The \emph{arm position} -- the main feature differentiating  Peaky (inward) and Stretchy (outwards).
\item The \emph{color of the animal} and the \emph{color of the background}, each defined by a scalar and a corresponding colormap.
\item The \emph{shape of the blocks} which continuously varies from spherical to cubic.
\item The \emph{deformation of the animal} which ``bends'' the animal's spine.
\item The \emph{3 degrees of freedom (DoF) rotation} and \emph{2 DoF positioning} of the animal in the image.
\end{itemize}

Our library samples parameters randomly. We provide a basic sampler and some specialized samplers that can be subclassed to produce customized datasets.
The resulting images, along with a segmentation mask are rendered using the Blender Python API. 

Although the arm position is the main discriminative feature for either Peaky or Stretchy,
we sample it from slightly overlapping distributions.
By increasing the overlap, the classification difficulty is increased, ensuring that biases predictive of a class will be used.
Biases can be introduced by correlating different parameters with an animal type.
For example, \emph{Peakies could be more frequently blue than Stretchies}. The Appendix, shows a code snippet that creates a rotation bias (See Listing \ref{lst:rotation_bias}).
Even more complex dependencies are possible, e.g., the animal's color is only predictive if rotated above 
a certain amount
\footnote{Although the biases can be complicated, the images are still 
more simplistic than natural images allowing to train a model quickly (and even training generative models such as VAEs or GANs). Even free resources such as Google Colab can be used for training - see our example at \url{https://colab.research.google.com/github/mschuessler/two4two/blob/trainKerasExample/examples/two4two_leNet.ipynb}.}.
Changing parameters' ranges and distributions also allows for the creation of datasets
with different levels of variability, e.g a larger range of
possible rotations.

\newcommand{\tftsub}{0.3\textwidth}
\newcommand{\tftinc}{0.3\textwidth}
\begin{figure}[t]
    \centering
    \begin{subfigure}[t]{\tftsub}
        \includegraphics[width=\tftinc]{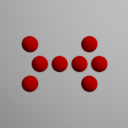}
        \includegraphics[width=\tftinc]{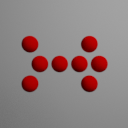}
        \includegraphics[width=\tftinc]{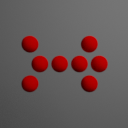}
        \caption{Background}
    \end{subfigure}
    \vspace{0.3cm}
    \begin{subfigure}[t]{\tftsub}
        \includegraphics[width=\tftinc]{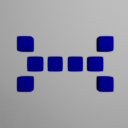}
        \includegraphics[width=\tftinc]{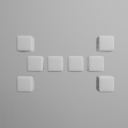}
        \includegraphics[width=\tftinc]{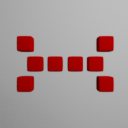}
        \caption{Color}
    \end{subfigure}

    \begin{subfigure}[t]{\tftsub}
        \includegraphics[width=\tftinc]{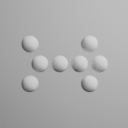}
        \includegraphics[width=\tftinc]{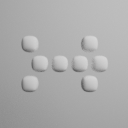}
        \includegraphics[width=\tftinc]{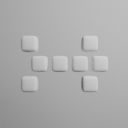}
        \caption{Shape (spheres or cubes)}
    \end{subfigure}
    \vspace{0.3cm}
    \begin{subfigure}[t]{\tftsub}
        \includegraphics[width=\tftinc]{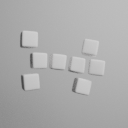}
        \includegraphics[width=\tftinc]{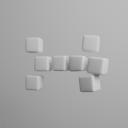}
        \includegraphics[width=\tftinc]{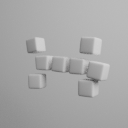}
        \caption{Rotation and Bending}
    \end{subfigure}
    \caption{Parameters in the \twofourtwo dataset.  
    Each subfigure, visualises possible changes to the object.
    \textbf{(a)} background and \textbf{(b)} animal colors can be changed.
    \textbf{(c)} The individual blocks can be spherical, cubic or something in between This is achieved by rounding off cubes until they become spherical. 
    \textbf{(d)} The animals can take a random pose.
    }
    \label{fig:two4two}
\end{figure}
\section{Human Subject Study}
We aimed to to validate whether we could use our library to generate data containing biases important to the model's predictions but are hard to notice for participants when solely inspecting the training data.

We noticed that many human subject studies in machine learning solely demonstrate their methods feasibility without a baseline comparison, e.g., \cite{Ribeiro2016,Singla2020}. Hence with our study design, we want to provide a reasonable baseline for further experiments.
A simple alternative to generated explanations are example-based explanations, in particular studying the system predictions on exemplary inputs.
This simple method has surfaced as a strong baseline in another study by \cite{borowski2020exemplary} and others studies already discovered that users predominantly rely on predictions rather than on generated explanations when reasoning about a model \citep{Chu2020,Adebayo2020}.
Therefore, we choose example-based explanations as our reference baseline treatment.

Considering that participants' attention is limited, we choose a 30x5 image grid (3 rows shown in Figure \ref{fig:user_study}).
Each column represented a classification score range.
We chose the ranges so that high confidence predictions for \emph{Stretchy} appeared on the far left column
and high confidence predictions \emph{Peaky} on the far right.
Less confident predictions were shown in the directly adjoining columns.
The remaining middle column represented borderline cases.
This visual design had prevailed throughout numerous iterations and ten pilot studies, as it allows users to quickly scan for similar features in columns and differing features in rows.

For our online study, we recruited 30 participants from Prolific.
The selection criteria was an academic degree with basic mathematical education.
Upon commencing the study on Qualtrics, we showed participants handcrafted video instructions about the study.
We asked them to study the model's output on the given images and form hypotheses about its used patterns and potential biases.
Concretely, we assessed participants' judgment about the plausibility of six hypotheses concerned with patterns the network learned on a 7-point Likert scale. Three hypotheses were reasonable (sensitivity to spatial compositions, color, and rotation). Two others were not (sensitivity to background and shape of individuals blocks).

\subsection{Study Results and Discussion}
\begin{figure}
    \centering
    \vspace{-1.20cm}
    \includegraphics[width=0.9\textwidth]{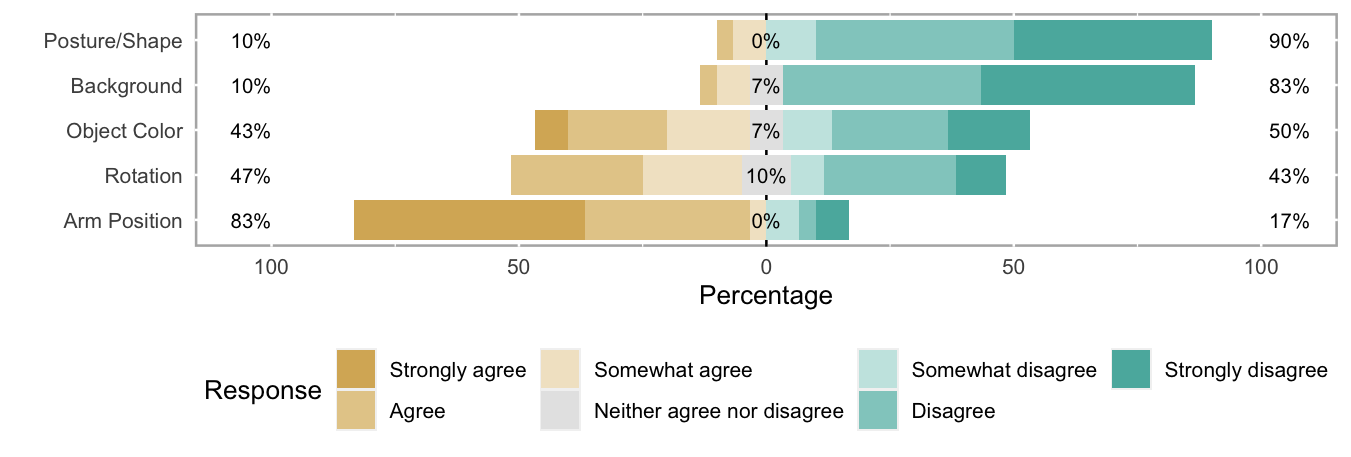}
    \vspace{-10pt}
    \caption{
    Participants agreements to statements about relevant patterns. 
    }
    \label{fig:user_study}
    \vspace{-0.5cm}
\end{figure}

Figure \ref{fig:user_study} summarizes the responses.
Most participants (83\%) correctly confirmed that the model is using the arm position for its prediction.
The color bias was identified by 43\% of the participants, and the rotation bias was identified by 47\% of the participants.
Irrelevant patterns were both correctly rejected by 90\% (sensitivity to posture) and 83\% (sensitivity to the background) of the participants.

We argue these results indicate a high quality of collected responses. We had succeeded in introducing two patterns in the dataset that were relevant enough to bias the system while only being noticeable to roughly half of the participants.

\section{Conclusion and Future Work}
In this work, we present the \twofourtwo synthetic dataset. Compared to BAM and CLEVER-XAI, we specifically designed it for human-subject evaluations.
We carefully crafted a simple scenario of two abstract animals, eliminating prior knowledge and beliefs as confounding factors in the experimental design.
At the same time, we provide extensive control over the data by a set of parameter distributions.
With our approach, researchers can either choose from a collection of pre-generated datasets or use our Python library to craft their own specific datasets by simply changing a few parameter distributions.
In the future, we will complement all pre-generated data sets with a user study scenario, including the metaphors, task instructions, and quantitative measures, ready to be used to instruct participants in crowdsourcing platforms.

As a first step towards this goal, we created a simple example-based baseline explanation technique and generated dataset with two biases.
In a user study (N=30), we confirmed that these biases were subtle enough to be noticed by only roughly half of the participants with the baseline method. Future studies can already use this study design to test whether they can beat this baseline with their explanation techniques.

With our work, we seek to facilitate a more rigorous science of interpretability by significantly lowering the barrier to conduct human subject evaluation of explanation techniques for image classification.

\newpage
\bibliography{iclr2021_conference}

\begin{thebibliography}{20}
\providecommand{\natexlab}[1]{#1}
\providecommand{\url}[1]{\texttt{#1}}
\expandafter\ifx\csname urlstyle\endcsname\relax
  \providecommand{\doi}[1]{doi: #1}\else
  \providecommand{\doi}{doi: \begingroup \urlstyle{rm}\Url}\fi

\bibitem[{Adadi} \& {Berrada}(2018){Adadi} and {Berrada}]{Adadi2018}
A.~{Adadi} and M.~{Berrada}.
\newblock Peeking inside the black-box: A survey on explainable artificial
  intelligence (xai).
\newblock \emph{IEEE Access}, 6:\penalty0 52138--52160, 2018.
\newblock ISSN 2169-3536.
\newblock \doi{10.1109/ACCESS.2018.2870052}.

\bibitem[Adebayo et~al.(2018)Adebayo, Gilmer, Muelly, Goodfellow, Hardt, and
  Kim]{adebayo2018sanity}
Julius Adebayo, Justin Gilmer, Michael Muelly, Ian Goodfellow, Moritz Hardt,
  and Been Kim.
\newblock Sanity checks for saliency maps.
\newblock \emph{Advances in Neural Information Processing Systems},
  31:\penalty0 9505--9515, 2018.

\bibitem[Adebayo et~al.(2020)Adebayo, Muelly, Liccardi, and Kim]{Adebayo2020}
Julius Adebayo, Michael Muelly, Ilaria Liccardi, and Been Kim.
\newblock Debugging tests for model explanations, 2020.

\bibitem[Alqaraawi et~al.(2020)Alqaraawi, Schuessler, Weiß, Costanza, and
  Berthouze]{Alqaraawi2020}
Ahmed Alqaraawi, Martin Schuessler, Philipp Weiß, Enrico Costanza, and Nadia
  Berthouze.
\newblock Evaluating saliency map explanations for convolutional neural
  networks: A user study.
\newblock In \emph{Proceedings of the 25th International Conference on
  Intelligent User Interfaces}, IUI ’20, pp.\  263–274, New York, NY, USA,
  2020. Association for Computing Machinery.
\newblock \doi{10.1145/3377325.3377519}.

\bibitem[Arras et~al.(2020)Arras, Osman, and Samek]{arras2021}
Leila Arras, Ahmed Osman, and Wojciech Samek.
\newblock Ground truth evaluation of neural network explanations with
  clevr-xai, 2020.

\bibitem[Borowski et~al.(2020)Borowski, Zimmermann, Schepers, Geirhos, Wallis,
  Bethge, and Brendel]{borowski2020exemplary}
Judy Borowski, Roland~S. Zimmermann, Judith Schepers, Robert Geirhos, Thomas
  S.~A. Wallis, Matthias Bethge, and Wieland Brendel.
\newblock Exemplary natural images explain cnn activations better than feature
  visualizations, 2020.

\bibitem[Chu et~al.(2020)Chu, Roy, and Andreas]{Chu2020}
Eric Chu, Deb Roy, and Jacob Andreas.
\newblock Are visual explanations useful? a case study in model-in-the-loop
  prediction, 2020.

\bibitem[Doshi-Velez \& Kim(2017)Doshi-Velez and Kim]{Doshi2017}
Finale Doshi-Velez and Been Kim.
\newblock Towards a rigorous science of interpretable machine learning.
\newblock \emph{arXiv: 1702.08608}, 2017.

\bibitem[Goyal et~al.(2019)Goyal, Wu, Ernst, Batra, Parikh, and Lee]{Goyal2019}
Yash Goyal, Z.~Wu, J.~Ernst, Dhruv Batra, D.~Parikh, and Stefan Lee.
\newblock Counterfactual visual explanations.
\newblock \emph{ArXiv}, abs/1904.07451, 2019.

\bibitem[Kaur et~al.(2020)Kaur, Nori, Jenkins, Caruana, Wallach, and
  Wortman~Vaughan]{Kaur2020}
Harmanpreet Kaur, Harsha Nori, Samuel Jenkins, Rich Caruana, Hanna Wallach, and
  Jennifer Wortman~Vaughan.
\newblock Interpreting interpretability: Understanding data scientists' use of
  interpretability tools for machine learning.
\newblock In \emph{Proceedings of the 2020 CHI Conference on Human Factors in
  Computing Systems}, CHI '20, pp.\  1–14, New York, NY, USA, 2020.
  Association for Computing Machinery.
\newblock ISBN 9781450367080.
\newblock \doi{10.1145/3313831.3376219}.
\newblock URL \url{https://doi.org/10.1145/3313831.3376219}.

\bibitem[Kim et~al.(2016)Kim, Khanna, and Koyejo]{Kim2016}
Been Kim, Rajiv Khanna, and Oluwasanmi~O Koyejo.
\newblock Examples are not enough, learn to criticize! criticism for
  interpretability.
\newblock In D.~Lee, M.~Sugiyama, U.~Luxburg, I.~Guyon, and R.~Garnett (eds.),
  \emph{Advances in Neural Information Processing Systems}, volume~29. Curran
  Associates, Inc., 2016.
\newblock URL
  \url{https://proceedings.neurips.cc/paper/2016/file/5680522b8e2bb01943234bce7bf84534-Paper.pdf}.

\bibitem[Leavitt \& Morcos(2020)Leavitt and Morcos]{leavitt2020towards}
Matthew~L Leavitt and Ari Morcos.
\newblock Towards falsifiable interpretability research.
\newblock \emph{arXiv preprint arXiv:2010.12016}, 2020.

\bibitem[Nie et~al.(2018)Nie, Zhang, and Patel]{nie2018theoretical}
Weili Nie, Yang Zhang, and Ankit Patel.
\newblock A theoretical explanation for perplexing behaviors of
  backpropagation-based visualizations.
\newblock In \emph{International Conference on Machine Learning}, pp.\
  3809--3818. PMLR, 2018.

\bibitem[Nunes \& Jannach(2017)Nunes and Jannach]{Nunes2017}
Ingrid Nunes and Dietmar Jannach.
\newblock A systematic review and taxonomy of explanations in decision support
  and recommender systems.
\newblock \emph{User Modeling and User-Adapted Interaction}, 27\penalty0
  (3-5):\penalty0 393--444, December 2017.
\newblock ISSN 0924-1868.
\newblock \doi{10.1007/s11257-017-9195-0}.
\newblock URL \url{https://doi.org/10.1007/s11257-017-9195-0}.

\bibitem[Ribeiro et~al.(2016)Ribeiro, Singh, and Guestrin]{Ribeiro2016}
Marco~Tulio Ribeiro, Sameer Singh, and Carlos Guestrin.
\newblock "why should i trust you?": Explaining the predictions of any
  classifier.
\newblock \emph{Proceedings of the 22nd ACM SIGKDD International Conference on
  Knowledge Discovery and Data Mining}, 2016.

\bibitem[Shen \& Huang(2020)Shen and Huang]{Shen2020}
Hua Shen and Ting-Hao~Kenneth Huang.
\newblock How {{Useful Are}} the {{Machine}}-{{Generated Interpretations}} to
  {{General Users}}? {{A Human Evaluation}} on {{Guessing}} the {{Incorrectly
  Predicted Labels}}.
\newblock In \emph{Proceedings of the {{Eighth AAAI Conference}} on {{Human
  Computation}} and {{Crowdsourcing}} ({{HCOMP}}-20)}, volume~8, pp.\
  168--172, {Virtual}, October 2020. {AAAI Press}.
\newblock ISBN 978-1-57735-848-0.

\bibitem[Singla et~al.(2020)Singla, Pollack, Chen, and
  Batmanghelich]{Singla2020}
Sumedha Singla, Brian Pollack, Junxiang Chen, and Kayhan Batmanghelich.
\newblock Explanation by progressive exaggeration.
\newblock In \emph{International Conference on Learning Representations}, 2020.

\bibitem[Sixt et~al.(2020)Sixt, Granz, and Landgraf]{sixt2020wel}
Leon Sixt, Maximilian Granz, and Tim Landgraf.
\newblock When explanations lie: Why many modified {BP} attributions fail.
\newblock In Hal~Daumé III and Aarti Singh (eds.), \emph{Proceedings of the
  37th International Conference on Machine Learning}, volume 119 of
  \emph{Proceedings of Machine Learning Research}, pp.\  9046--9057. PMLR,
  13--18 Jul 2020.
\newblock URL \url{http://proceedings.mlr.press/v119/sixt20a.html}.

\bibitem[Wortman~Vaughan \& Wallach(2021)Wortman~Vaughan and
  Wallach]{wortmanvaughan2021a}
Jennifer Wortman~Vaughan and Hanna Wallach.
\newblock A human-centered agenda for intelligible machine learning.
\newblock This is a draft version of a chapter in a book to be published in the
  2020 - 21 timeframe., February 2021.
\newblock URL
  \url{https://www.microsoft.com/en-us/research/publication/a-human-centered-agenda-for-intelligible-machine-learning/}.

\bibitem[Yang \& Kim(2019)Yang and Kim]{Yang2019}
Mengjiao Yang and Been Kim.
\newblock Benchmarking attribution methods with relative feature importance,
  2019.

\end{thebibliography}
\bibliographystyle{iclr2021_conference}

\appendix

\definecolor{mygreen}{rgb}{0,0.6,0}
\definecolor{mygray}{rgb}{0.5,0.5,0.5}
\definecolor{mymauve}{rgb}{0.58,0,0.82}

\lstset{ 
  backgroundcolor=\color{white},   
  basicstyle=\footnotesize,        
  breakatwhitespace=false,         
  breaklines=true,                 
  captionpos=b,                    
  commentstyle=\color{mygreen},    
  deletekeywords={...},            
  escapeinside={\%*}{*)},          
  extendedchars=true,              
  firstnumber=1,                
  frame=single,	                   
  keepspaces=true,                 
  keywordstyle=\color{blue},       
  language=Octave,                 
  morekeywords={*,...},            
  numbers=left,                    
  numbersep=5pt,                   
  numberstyle=\tiny\color{mygray}, 
  rulecolor=\color{black},         
  showspaces=false,                
  showstringspaces=false,          
  showtabs=false,                  
  stepnumber=1,                    
  stringstyle=\color{mymauve},     
  tabsize=2,	                   
  title=\lstname                   
}

\vspace{1cm}
\begin{lstlisting}[
    label={lst:rotation_bias}, 
    language=Python, 
    caption=Source code example to create a biased sampler. High positive rotations are 
    predictive of Stretchy and low negative rotations of Peaky.
    ]
import dataclasses
import numpy as np
import matplotlib.pyplot as plt

from two4two.blender import render
from two4two.bias import Sampler, Continouos
from two4two.scene_parameters import SceneParameters

@dataclasses.dataclass
class RotationBiasSampler(Sampler):
    """A rotation-biased sampler.
    
    The rotation is sampled conditionally depending on the object type.
    Positive rotations for peaky and negative rotations for stretchy.
    """
    
    obj_rotation_yaw: Continouos = dataclasses.field(
        default_factory=lambda: {
            'peaky': np.random.uniform(-np.pi / 4, 0),
            'stretchy': np.random.uniform(0, np.pi / 4),
        })

# sample a 4 images
sampler = RotationBiasSampler()
params = [sampler.sample() for _ in range(4)]
for img, mask, param in render(params):
    plt.imshow(img)
    plt.title(f"{param.obj_name}: {param.obj_rotation_yaw}")
    plt.show() 
\end{lstlisting}

\end{document}